\documentclass[10pt,twocolumn,letterpaper]{article}

\usepackage{iccv}
\usepackage{times}
\usepackage{epsfig}
\usepackage{graphicx}
\usepackage{amsmath}
\usepackage{amssymb}
\usepackage[toc,page]{appendix}

\usepackage{xcolor}

\usepackage[breaklinks=true,bookmarks=false]{hyperref}

\iccvfinalcopy 

\def\iccvPaperID{****} 

\ificcvfinal\fi

\begin{document}

\title{Blocks World Revisited: \\The Effect of Self-Occlusion on Classification by Convolutional Neural Networks}

\author{Markus D. Solbach\\
York University\\
{\tt\small solbach@eecs.yorku.ca}
\and
John K. Tsotsos\\
York University\\
{\tt\small tsotsos@eecs.yorku.ca}
}

\maketitle

\begin{abstract}
   Despite the recent successes in computer vision, there remain new avenues to explore. 
   In this work, we propose a new dataset to investigate the effect of self-occlusion on deep neural networks. 
   With \textit{TEOS} (The Effect of Self-Occlusion), we propose a 3D blocks world dataset that focuses on the geometric shape of 3D objects and their omnipresent challenge of self-occlusion. We designed \textit{TEOS} to investigate the role of self-occlusion in the context of object classification. Even though remarkable progress has been seen in object classification, self-occlusion is a challenge. In the real-world, self-occlusion of 3D objects still presents significant challenges for deep learning approaches. However, humans deal with this by deploying complex strategies, for instance, by changing the viewpoint or manipulating the scene to gather necessary information. 
   With \textit{TEOS}, we present a dataset of two difficulty levels ($L_1$ and $L_2$), containing 36 and 12 objects, respectively. We provide 738 uniformly sampled views of each object, their mask, object and camera position, orientation, amount of self-occlusion, as well as the CAD model of each object. We present baseline evaluations with five well-known classification deep neural networks and show that \textit{TEOS} pose a significant challenge for all of them. The dataset, as well as the pre-trained models, are made publicly available for the scientific community under \url{https://data.nvision.eecs.yorku.ca/TEOS/}. 
\end{abstract}

\section{Introduction}

Over most of the last decade, computer vision was pushed by efforts put into deep learning. The exact advent of this deep learning dominated era is often dated to the ImageNet challenge (\cite{ILSVRC15}) in 2012. Since then, the performance of models on various tasks has been improving at unparalleled speed; for instance, image classification on the ImageNet dataset surpassed the reported human-level performance in 2015 (\cite{He2015}). Two of the enablers for the recent successes are faster computers, specifically graphic processors, and the availability of large scale and often well-curated data sets to learn from.  
\\The deep learning paradigm is omnipresent, and, with it, the need for data with specific statistics to work in certain domains. \cite{Kuznetsova2020} goes as far as saying that "Data is playing an especially critical role in enabling computers to interpret images as compositions of objects, an achievement that humans can do effortlessly while it has been elusive for machines so far." 
\\Many domains exist in which one would like machines to perform visual tasks (\cite{carroll1993human}). One of these is object classification, which is defined as whether a particular item is present in the stimulus (\cite{Tsotsos2005}).  
\\Object classification is an essential capability of humans, as well as for any robotic system whose goal is to be a real-world assistant; in a factory, hospital, or at home, just to name a few. Even though very successful in many domains, deep learning methods are challenged with occlusion (\cite{Koporec2019}), which is inevitable in real-world scenarios. Here, we go a step further and show that deep learning methods are also challenged with the self-occlusion of objects, hence not generalizing to objects' 3D structure.

\begin{figure}[t]
\begin{center}
   \includegraphics[width=1.0\linewidth]{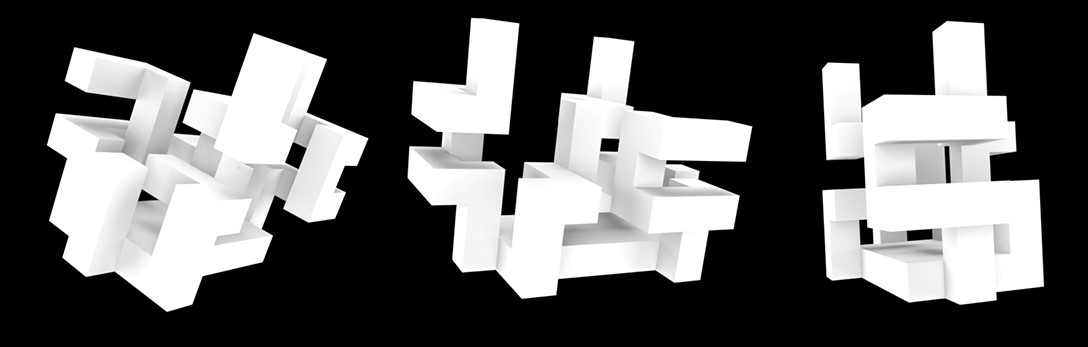}
\end{center}
   \caption{Example of the proposed objects from three different viewpoints.}
\label{fig:exampleobject}
\end{figure}

\begin{figure}[th!]
\begin{center}
   \includegraphics[width=1.0\linewidth]{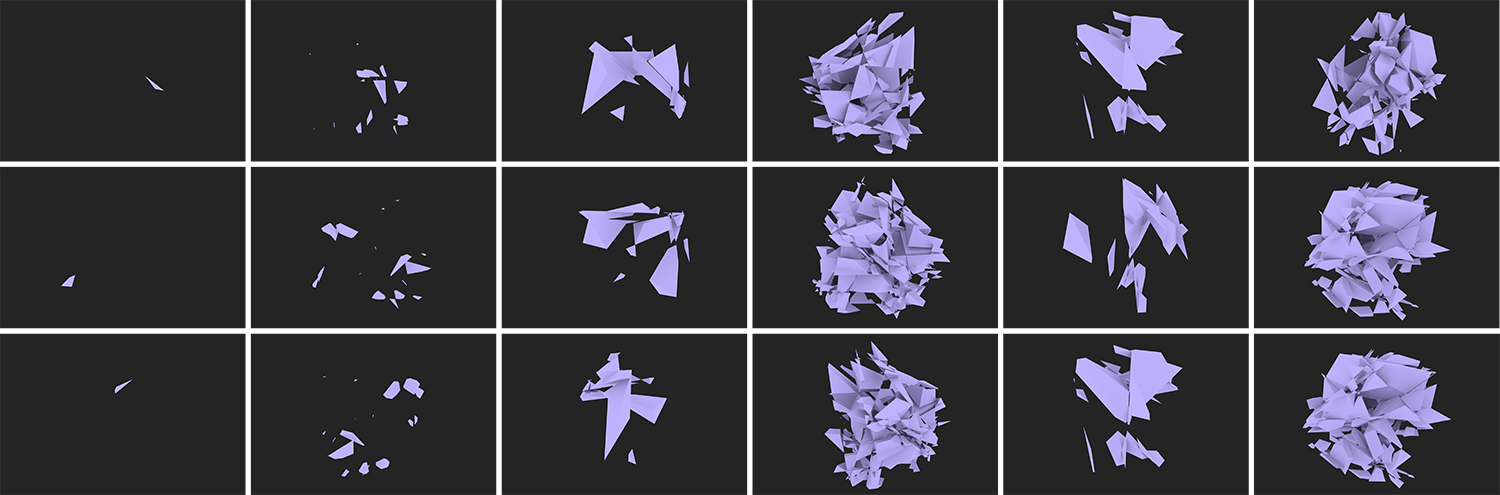}
\end{center}
   \caption{Six polyhedral scenes from three different viewpoints with increasing complexity (\cite{Solbach2018a}).}
\label{fig:psg}
\end{figure}

The problem of understanding the 3D structure from a 2D description, for instance, a line drawing, was first put forward independently by \cite{huffman1971impossible} and \cite{clowes1971seeing}, and they both showed that the necessary critical condition for a line drawing to represent an actual arrangement of polyhedral objects was labelability. 
\\As the human brain is very efficient at reconstructing a scene's 3D structure from a single image with no texture, colour or shading, efforts have been concentrated on computational complexity issues; one might think an efficient solution exists (e.g. polynomial-time). \cite{kirousis1988complexity}, however, proved that this problem is NP-Complete, also for simple cases like trihedral, solid scenes. To further research in this field, \cite{Parodi1998b} proposed a method to generate random instances of line drawings with useful distribution to investigate questions related to complexity of understanding images of polyhedral scenes. More recently, \cite{Solbach2018a} provided a 3D extension with controllable camera parameters and two different light settings. It is designed to enable research on how a program could parse a scene if it had multiple and definable viewpoints to consider. An example of a polyhedral scene from \cite{Solbach2018a} is shown from three different viewpoints in Figure \ref{fig:psg}. 

With the increasing successes, contemporary computer vision approaches show a healthy trend away from artificial problems and provide solutions to real-world problems, already deployed in many domains (\cite{Andreopoulos2013}), for example, optical character recognition, industrial inspection systems, medical imaging, and biometrics. However, at the same time, a disparagement of artificial domains can be seen (\cite{Slaney2001}). At the very least, these domains can support meaningful systematic experiments. Here we revisit one such artificial domain; the Blocks World.  
\\In visual perception, the basic physical and geometric constraints of our world play a crucial role. This idea goes back at least to Helmholtz and his argument for \textit{unconscious inference}.
\\This theme of visual perception can be traced back to the early years of the discipline of computer vision. Larry Roberts argued that "the perception of solid objects is a process which can be based on the properties of three-dimensional transformations and the laws of nature" (\cite{Roberts1965}). Roberts' popular Blocks World was an early attempt to build a system for complete scene understanding for a closed artificial world of textureless polyhedral shapes by using a generic library of polyhedral block shapes. This toy domain that has remained as a staple of the AI literature for over 30 years but has fallen into disrepute since then. This is due to a superficial understanding of it, leading to insufficient experimental methodology, and therefore, failing to extract useful results from it (\cite{Slaney2001}). 

\begin{figure}[th!]
\begin{center}
   \includegraphics[width=0.5\linewidth]{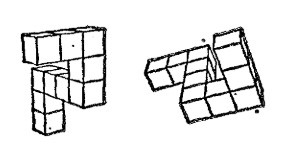}
\end{center}
   \caption{Example of the objects by \cite{Shepard2019} which are used as an inspiration.}
\label{fig:shepard}
\end{figure}

In this paper, we present \textit{TEOS}: The Effect of Self-Occlusion. \textit{TEOS} is a Blocks World based set of objects with known complexity, controlled viewpoints, with a known level of self-occlusion and 3D models. \textit{TEOS} shares similarities in appearance with the so-called Shepard and Metzler Objects (\cite{Shepard2019}), which are widely used in the literature for mental rotation tasks. See Figure \ref{fig:shepard} for an illustration of two such objects. Similarities are for instance, the strict ninety-degree angle of elements making up an object, the use of only cuboids, the use of mainly one primitive (except for the base plate). 
\\However, with \textit{TEOS}, we present a set of objects that go beyond the Shepard and Metzler objects. Specifically, our objects have a known complexity, share a common coordinate system, and empirical results have shown that they are challenging for visual tasks using state-of-the-art classification algorithms.
\\Our contributions are an investigation of the effect of self-occlusion for object classification. To accomplish this, we provide a novel set of objects, a carefully created dataset, including an in-depth explanation of the objects and generated data with a focus on self-occlusion and a baseline evaluation with state-of-the-art classification algorithms. 
\\The remainder of the paper is structured as follows. First, we will explain in detail the objects we have created for \textit{TEOS}. We then continue by giving an overview of related work, describing the data acquisition, presenting our self-occlusion measure, evaluating the dataset against state-of-the-art classification algorithms, and finally finishing with our conclusions and future directions. 

\section{Related Work}

To the best of our knowledge, self-occlusion has not attracted much attention in the literature. However, occlusion caused by other objects has. In addition to several datasets, a number of approaches were introduced to deal with occlusion.

\subsection{Occlusion Datasets}

\begin{figure}[th!]
\begin{center}
   \includegraphics[width=0.8\linewidth]{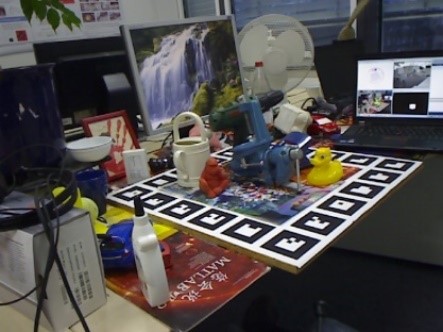}
\end{center}
   \caption{A scene with different objects under occlusion taken from the \textit{ICCV 2015 Occluded Object Challenge}.}
\label{fig:iccv2015}
\end{figure}

A burden of deep learning is its need for vast mounts of training data. Even though occlusion and its effect on vision tasks has been addressed for some time (\cite{Hsiao2010, Ouyang2012, Brachmann2014, Hsiao2014}), datasets created are usually too small to be used to train successful deep learning models. Furthermore, to our knowledge, datasets, if considering occlusion, mostly introduce various levels of clutter but lack to define occlusion in a generic way.  
For instance, the CMU Kitchen Occlusion dataset (CMU\_KO8) by \cite{Hsiao2014} consists of 1,600 images of eight kitchen objects, such as backing pan, scissors, etc., which only yields 200 examples per class. The dataset has explicitly been designed to challenge object recognition algorithms with strong viewpoint and illumination changes, occlusions and clutter. Besides a novel dataset, an occlusion reasoning module is also proposed (further details in Section \ref{sec:occreasoning}). 
\\With the \textit{ICCV 2015 Occluded Object Challenge} (\cite{Hinterstoisser2013a, Brachmann2014}), a dataset with eight objects positioned in a realistic setting of heavy occlusion is presented. The objects can be described as being of different domains, ranging from animals, over office supplies, to kitchenware. However, neither a definition of occlusion nor a metric is given. Figure \ref{fig:iccv2015} shows an example image of the dataset. 

The majority of occlusion datasets, however, deal with the occlusion of pedestrians. Specifically, in the context of autonomous driving, detecting pedestrians, even if occluded, is crucial to detect potential collisions. It is argued that most existing datasets are not designed for evaluating occlusion. For instance, the Caltech dataset (\cite{Dollar2012}) only contains 105 out of 4250 images with occluded pedestrians. The CUHK Occlusion Dataset (\cite{Ouyang2012}) is specifically designed as a pedestrian dataset with occlusion. The authors selected images from popular pedestrian datasets and recorded images from surveillance cameras and filtered them mainly for occluded pedestrians. The dataset contains 1,063 images with binary classification to indicate whether the pedestrian is occluded or not.  

\subsection{Occlusion Reasoning}\label{sec:occreasoning}

Reasoning about occlusion has been used in many areas, from object recognition to tracking and segmentation. Reported in (\cite{Hsiao2014}, the literature is extensive, but there has been comparatively little work on modelling occlusions from different viewpoints and using 3D information until recently. Further, occlusion reasoning is broadly classified into five categories; inconsistent object statistics, multiple images, part-based models, 3D reasoning,  and convolutional neural networks.  
\\The first category includes classical approaches, which use inconsistent object statistics to reason about potential occlusion. In general, occlusions are modelled as regions that are inconsistent with object statistics. For instance, \cite{meger2011explicit} use inconsistencies in 3D sensor data to classify occlusions. \cite{girshick2011object} introduce an occluder part in their grammar model when all parts cannot be placed. \cite{wang2009hog} use a scoring metric based on individual HOG filter cells. \cite{Hsiao2014} incorporate occlusion reasoning in object detection in a two-stage manner. First, in a bottom-up stage, occluded regions are hypothesized from image data. Second, a top-down stage is used that relies on prior knowledge to score the candidates' occlusion plausibility. Extensive evaluation on single, as well as multiple views show that incorporating occlusion reasoning yields significant improvement in recognizing texture-less objects under severe occlusions. 
\\The use of multiple images characterizes the second category. For these approaches, a sequence of consecutive images is necessary to disambiguate the object from occluders. For instance, \cite{ess2009improved} detects the objects and extrapolates the state of occluded objects using an Extended Kalman Filter. Reliable tracklets that are used in a temporal sliding window fashion are generated to disambiguate occluded objects in \cite{xing2009multi}.  
\\One of the largest categories are approaches using part-based models. A challenge of global object templates is occlusion as their performance degrades with its presence significantly. A popular solution to this problem is to separate the object into a set of parts and detect parts individually. This approach generally yields more robust detections towards occlusion. For example, \cite{shu2012part} analyze the contribution of each part using a linear SVM and train the classifier to use unoccluded parts to maximize the probability of detection. \cite{wu2009detection} go a step further and use multiple part detectors to maximize the joint likelihood. Binary classification of parts is introduced by \cite{vedaldi2009structured}. They decompose the HOG descriptor into small blocks that selectively switch between an object and an occlusion descriptor. 

\begin{figure}[th!]
\begin{center}
   \includegraphics[width=1.0\linewidth]{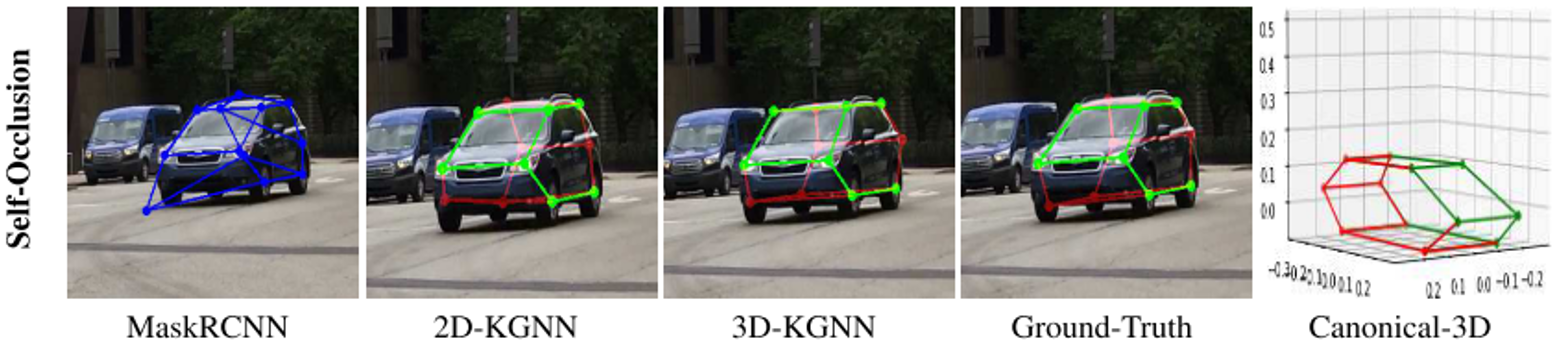}
\end{center}
   \caption{The effect of occlusion reasoning used in a CNN. Left the original CNN (MaskRCNN) and different (2D and 3D) occlusion reasoning approaches improve the detection (\cite{reddy2019occlusion}).}
\label{fg:red}
\end{figure}

More recent work using 3D information for occlusion reasoning has been introduced. Hsiao et al. argue that having 3D data provides richer information of the world, such as depth discontinuities and object size. \cite{pepikj2013occlusion} train multiple occlusion detectors on mined 3D annotated urban street scenes that contain distinctive, reoccurring occlusion patterns. \cite{wang2013learning} use RGB-D information and an extended Hough voting to include object location and its visibility pattern. \cite{Radwan2013} addresses precisely the problem of self-occlusion in the context of human pose estimation and adds an inference step to handle self-occlusion to an off-the-shelf body pose detector to increase its performance under self-occlusion. The solution leverages prior knowledge about the kinematics and orientations of the human pose to deal with self-occluding body parts.
\\Lastly, convolutional neural networks have shown promising results when it comes to different visual tasks like object classification, recognition, segmentation and 3D pose estimation. However, occlusion and, as will be shown in this work, self-occlusion pose significant challenges. \cite{reddy2019occlusion} introduce a framework to predict 2D and 3D locations of occluded key points for objects to mitigate the effect of occlusion on the performance. Evaluated on CAD data and a large image set of vehicles at busy city intersections, the approach increases the localization accuracy of MaskRCNN by about 10\%. A self-occlusion example can be seen in Figure \ref{fg:red}. \cite{Li2019a} uses deep supervision to fine-grain image classification. In their approach, they simulate challenging occlusion configurations between objects to enable reliable data-driven occlusion reasoning. Occlusion is modelled by rendering multiple object configurations and extracting the visibility level of the object of interest. With the work of \cite{Kortylewski2020}) a convolutional neural network is combined with the idea of part-based models. The authors introduce CompositionalNets. It is proposed to replace the fully-connected classification layer with a differentiable compositional model. The idea of CompositionalNets is to decompose images into objects and context, and then decompose objects into parts and objects' pose. Using various CNN backbone shows that the approach can learn features that are invariant to occlusion and discard occluders during classification, hence increasing performance, especially under occlusion. However, a trade-off is pointed out that a good occluder localization lowers classification performance because classification benefits from features that are invariant to occlusion, where, on the other hand, occluder localization requires a different type of features. Namely, ones that sensitive to occlusion. It is pointed out that it is essential to resolve this trade-off with new types of models. 

\section{Object Definitions}

In this Section, we will talk about the creation of the objects, as well as their characterization. 

With \textit{TEOS}, we present in total 48 objects, split into two sets; $L_1$ and $L_2$. The idea of having two sets is to support the different needs of research. $L_1$ consists of 36 objects in 18 complexity levels, hence tailored towards research exploring the effect of finely grained complexity changes. $L_2$, on the other side, is made up of 12 objects in three complexity levels based on $L_1$, so made up of complexity groups.

\begin{figure}[th!]
\begin{center}
   \includegraphics[width=0.8\linewidth]{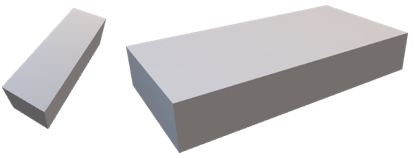}
\end{center}
   \caption{The building blocks used to create the objects of \textit{TEOS}; cuboid (left) and base (right).}
\label{fig:blocks}
\end{figure}

In a number of empirical studies, we have studied the relationship between the number of elements per object and classification accuracy. The human performance of classification $L_1$ objects is reliable (accuracy of $>98\%$) for objects with up to seven elements. The classification gets uncertainty (89\%) with objects consisting of around ten elements. Finally, the classification gets challenging (57\%) with objects made of around 18 elements.
\\Based on these findings, we have created the $L_2$ set with three complexity levels; easy with seven elements, medium with ten elements, and hard with 18 elements. For each complexity level, we have constructed three more objects with the same amount of elements but changing the orientation of one of the elements to have four distinct objects for each level. 
We will now continue to explain the elements used to assemble the objects and describe the objects' typical characteristics. 

All objects consist of the following two elements:

\begin{itemize}
   \item One 20mm x 60mm x 120mm base (Figure \ref{fig:blocks} right) 
   \item $n$ 20mm x 20mm x 60mm cuboids (Figure \ref{fig:blocks} left)
\end{itemize}

\begin{figure}[th!]
\begin{center}
   \includegraphics[width=0.8\linewidth]{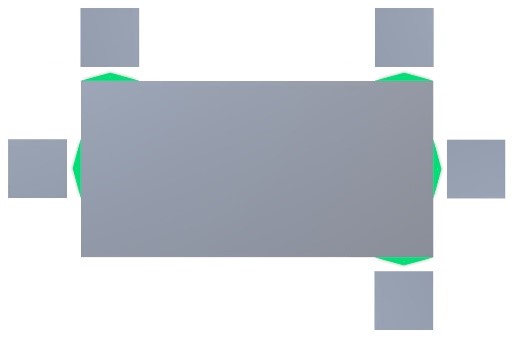}
\end{center}
   \caption{Possible connection points of cuboids on the base.}
\label{fig:connections}
\end{figure}

\begin{figure}[th!]
\begin{center}
   \includegraphics[width=0.5\linewidth]{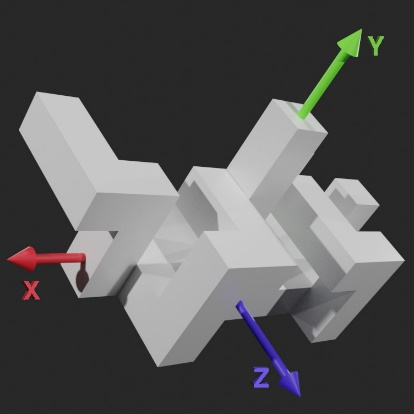}
\end{center}
   \caption{Illustration of the common coordinate system of the objects.}
\label{fig:coord}
\end{figure}

The complexity of an object is simply calculated as 

\begin{equation}
       compl = n + 1 
\end{equation}
Where $n$ is the number of cuboids used.
\\The base has five connection points for cuboids to be attached to. All cuboids are only attached upright, sitting flush with the bottom of the base. This also makes it simple to define a coordinate system. See Figure \ref{fig:coord} for an illustration.

All objects share the same coordinate system, which is crucial for any research that looks at the effect of the orientational difference of 3D objects. The coordinate system is defined as depicted in Figure \ref{fig:coord}; the Y-Axis is running orthogonal out of the base, the X-Axis running through the base from its center of gravity towards the end with three cuboids connectors, and the Z-Axis runs orthogonal to the Y- and X-Axis with the positive direction through the side of the base with two cuboid connections.

\begin{figure}[th!]
\begin{center}
   \includegraphics[width=0.6\linewidth]{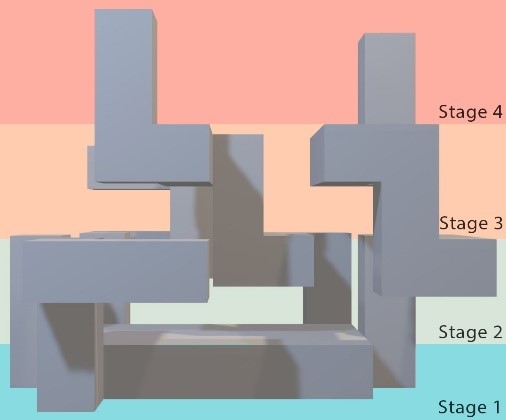}
\end{center}
   \caption{Stage-wise characterization of the objects. Another form of expressing the complexity of the object.}
\label{fig:stages}
\end{figure}

Building up the objects, a cuboid has eight connectors at which another cuboid can be attached. Consecutive cuboids are always orthogonally and never aligned in their direction, which is one of the differences to the Sheppard and Metzler objects. Furthermore, cuboids never intersect or touch neighbouring cuboids, hence avoiding geometrical loops. Creating the objects for $L_1$, we focused on making the complexity comparable by consecutively adding one cuboid per complexity level to the object of the previous complexity level.

\begin{figure}[th!]
\begin{center}
   \includegraphics[width=1.0\linewidth]{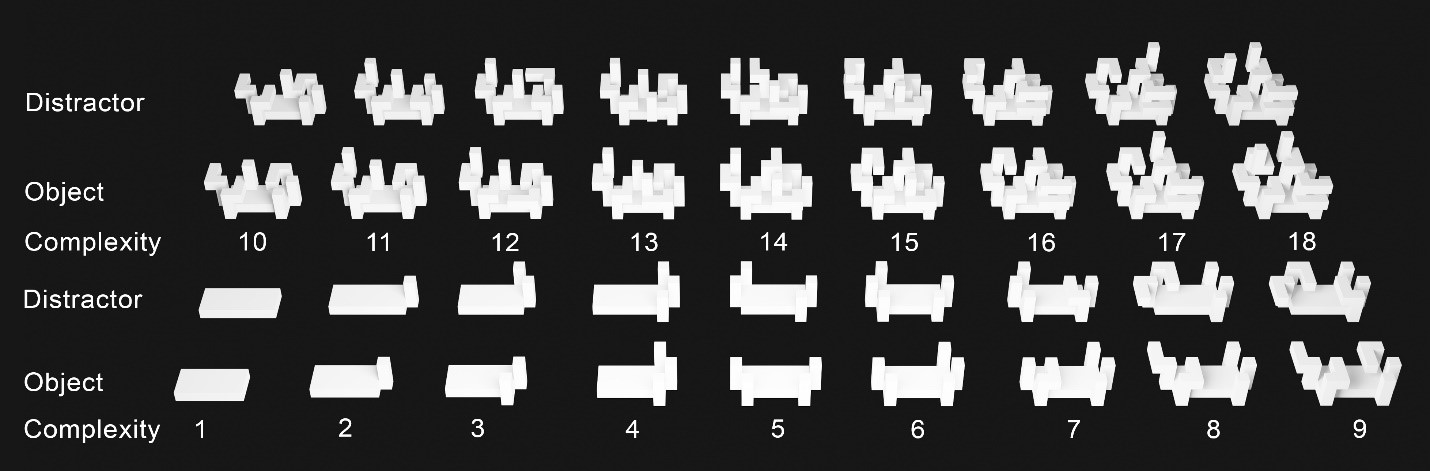}
\end{center}
   \caption{Illustration of $L_1$ with all 36 objects.}
\label{fig:l1}
\end{figure}

The objects can also be characterized in four stages of height. Each stage adds one perpendicular cuboid. For \textit{TEOS}, four height-stages were introduced. Consequently, the number of height-stages presents another way of expressing the complexity of an object. An illustration of the different stages is given in Figure \ref{fig:stages}.

\begin{figure}[th!]
\begin{center}
   \includegraphics[width=1.0\linewidth]{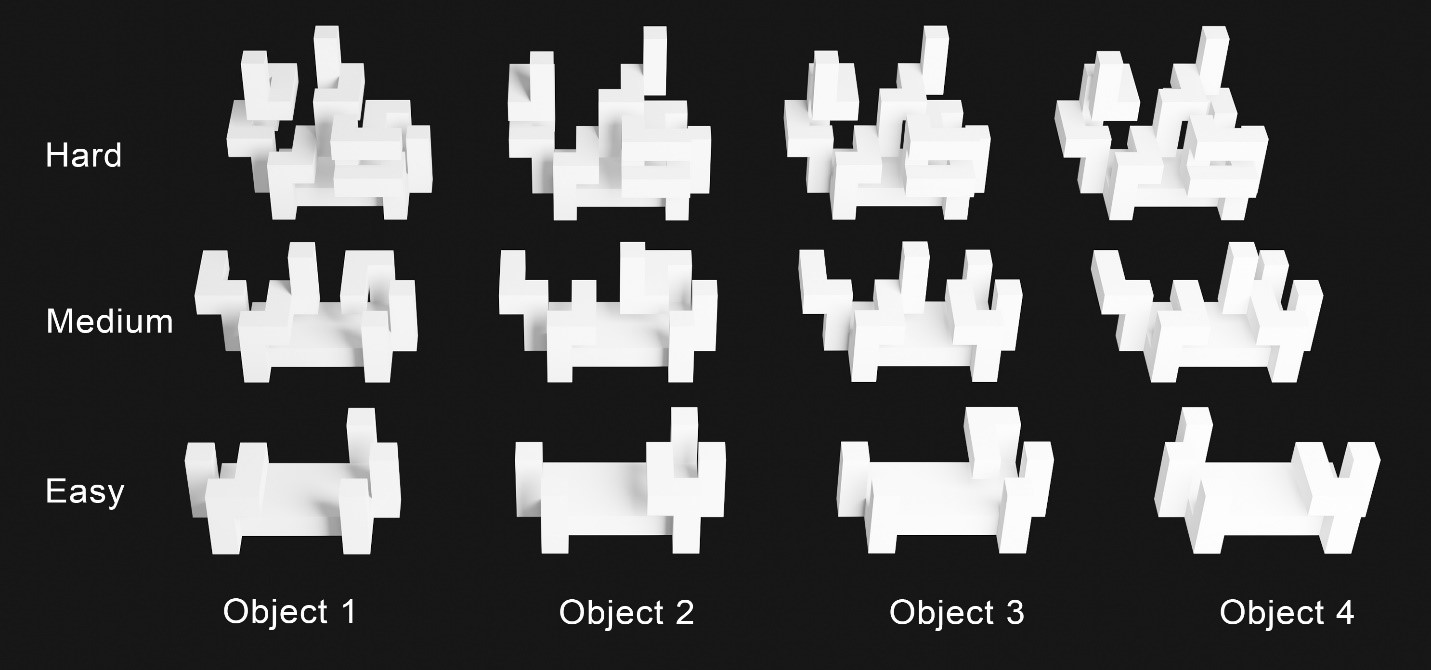}
\end{center}
   \caption{Illustration of $L_2$ with all 12 objects, split into three different complexity levels.}
\label{fig:l2}
\end{figure}

Lastly, having the objects characterized, we present the $L_1$ and $L_2$ object sets. The $L_1$ objects can be seen in Figure \ref{fig:l1}, and consist of 36 objects split into 18 complexity levels. There is a distractor object of the same complexity for each object that differs only in one small detail; one of the items is oriented differently. The introduction of the distractor objects is intended to support research in visual recognition, where merely counting the number of elements would reveal the object class.
\\On the other side, the $L_2$ set of objects is designed with less variation in complexity but more variation within a complexity. Twelve objects are evenly split into three complexity levels. To stick with the analogy of the distractors of the $L_1$ set, each complexity level of $L_2$ has three distractors; the same amount of elements but slight changes in assembling them. As described in the introduction of this Section, we chose the complexity levels based on empirical studies of visual recognition tasks. 
The $L_2$ objects can be seen in Figure \ref{fig:l2}.

\begin{figure}[th!]
\begin{center}
   \includegraphics[width=1.0\linewidth]{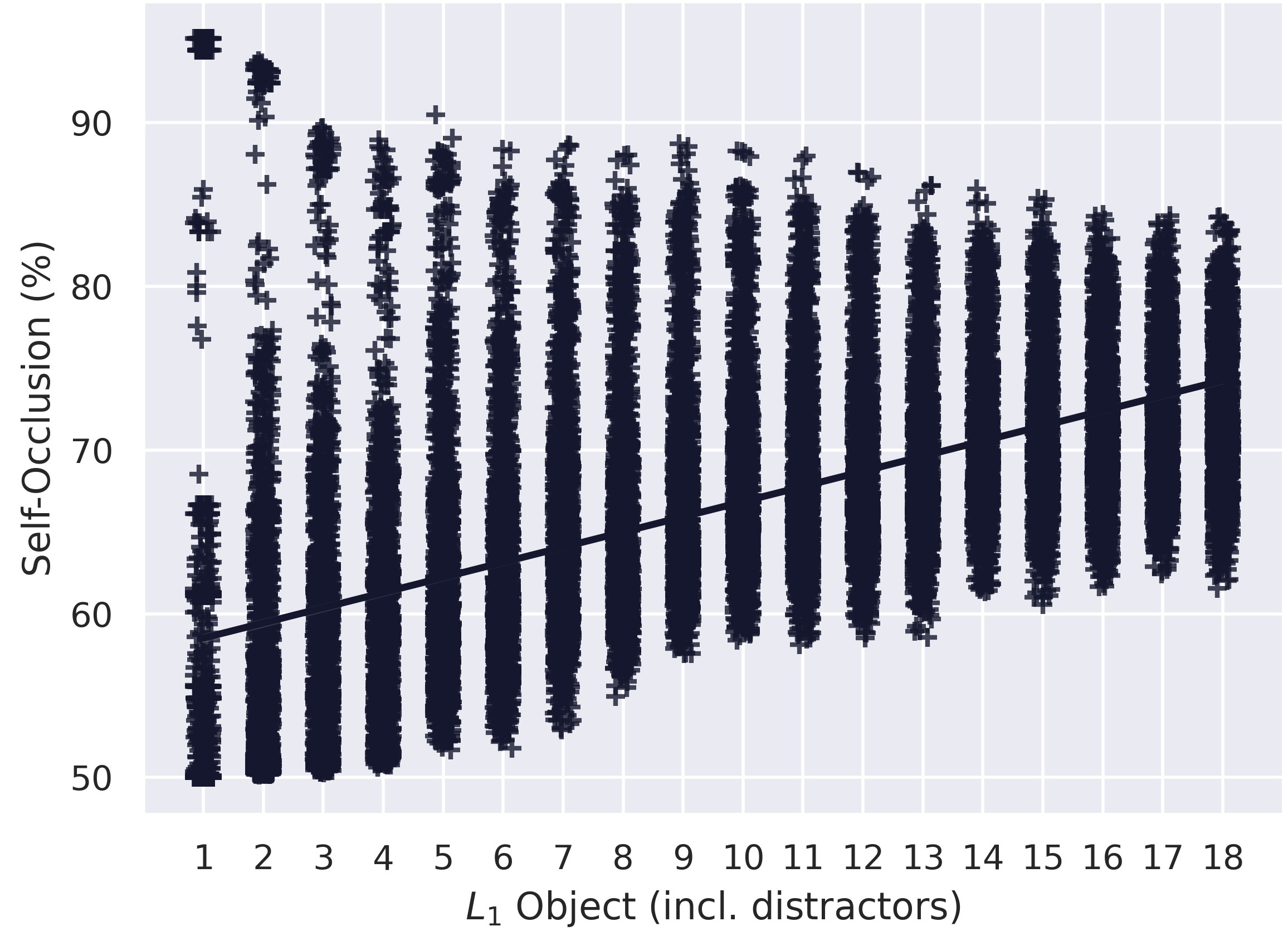}
\end{center}
   \caption{Illustration of the amount of average self-occlusion per object of $L_1$.}
\label{fig:occ_l1_dis}
\end{figure}

An emphasis in designing these datasets was put on the ability to use them for self-occlusion studies. To accomplish this, all objects are built around a common coordinate system, share the same base and progressively add elements to increase the complexity and level of self-occlusion. Figure \ref{fig:occ_l1_dis} shows how an increase of complexity of the $L_1$ dataset also increases the average amount of self-occlusion among all viewpoints. However, worth noting, with an increasing amount of complexity, the self-occlusion distribution per class decreases. Further information about our self-occlusion measure will be explained in Section \ref{sec:som}.

\section{Dataset Acquisition}

\textit{TEOS} is a dataset that is designed to be used in the virtual as well as the real world. For the former, one can use the rendered images and provided 3D Models (.STL file). For the latter, the objects are designed to be printable with a 3D printer. For this, we have prepared the 3D Models to be readable by many 3D printing slicing software. However, in this Section we want to focus on the generation of the rendered dataset images for which we have used Blender (\cite{Community2020}), a free and open-source 3D computer graphics software toolset. 
For \textit{TEOS}, each object was rendered from 768 views – Totalling 36,864 images. To achieve realistic renderings of the objects, we used the Cycles Path Tracing rendering engine, created a white, smooth, plastic imitating material, set six light sources in the rendering scene and used 4096 paths to trace each pixel in the final rendering.

\begin{figure}[th!]
\begin{center}
   \includegraphics[width=0.8\linewidth]{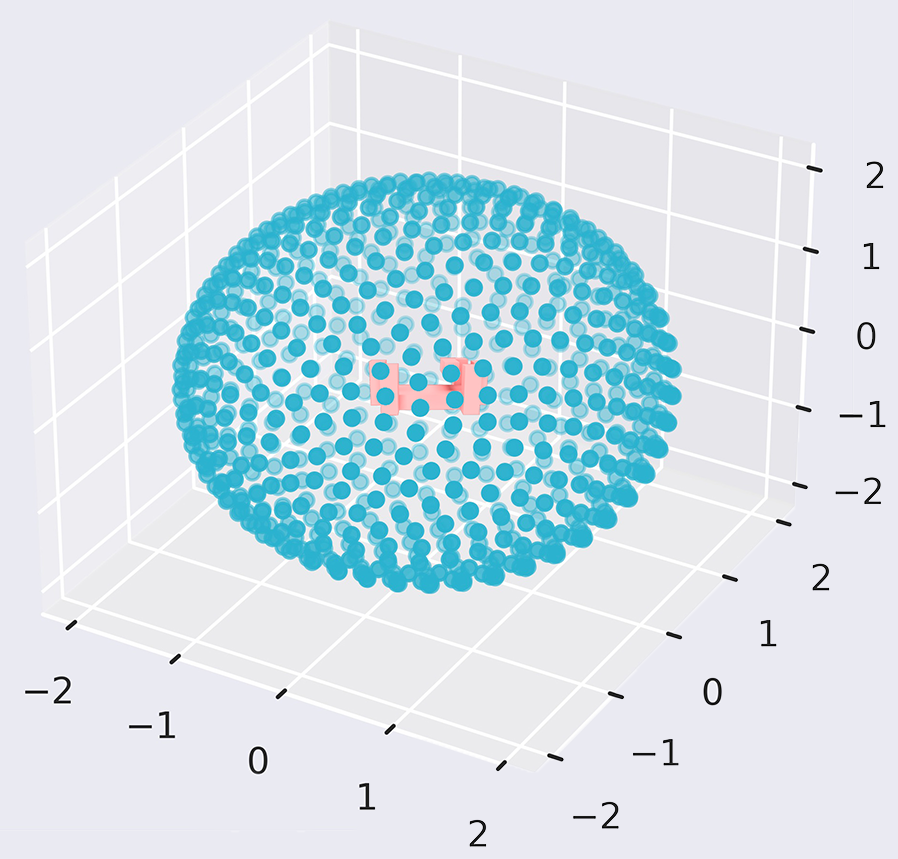}
\end{center}
   \caption{Illustration of viewpoints used to render each object of $L_1$ and $L_2$. Views are evenly distributed on a sphere around an object (blue points) and point towards the object (light red). In total 768 views are taken.}
\label{fig:views}
\end{figure}

Each object is rendered from the same set of views. To determine the views, we used the Fibonacci lattice (\cite{stanley1988differential}) approach. This approach allows distributing points on a sphere uniformly. Other approaches, for example, using radial distance, polar angle and azimuthal angle, will result in an unevenly sampled sphere; dense on the poles and sparse closer to the equator.  Figure \ref{fig:views} illustrates the chosen views to generate the dataset. Each blue-coloured point represents a location where the camera is placed and oriented to the center where the object (red) is. We chose a sphere radius of two such that the object is view-filling but not cropped in any view. 
\\Further, as it is sometimes practiced in the machine learning community (\cite{Everingham2010,Lin2014,dsprites17,multiobjectdatasets19}), we also provide the object mask and renderings with a dark and bright background for data augmentation purposes. 
The annotation file contains the object-type, view-id, bounding box information, object and camera positions and orientations, and object dimensions. 

\section{Self-Occlusion Measure}\label{sec:som}

It seems evident that if we see less of an object, it is harder to classify it. Regions of the object that are occluded to us might hold distinct features to tell object $X$ apart from object $Y$. In other words, occlusion for visual classification plays an important role. However, it is not only dependent on the view but also on the object. Let us take, for example, a sphere. No matter from which angle we look at it, we always observe 50\% of it. On the contrary, for a complex polygonal shape, this cannot be answered as quickly as it is dependent on its geometry. 
\\\cite{Gay-Bellile2010} distinguishes two kinds of occlusions; ``external occlusion'' and ``self-occlusion.'' ``External occlusion'' is caused by an object entering the space between the camera and the object of interest and ``self-occlusion ''which describes the occlusion caused by the object of interest to itself. For \textit{TEOS}, we are interested in the latter, as we always have one object in the scene. 
To our knowledge, no standard self-occlusion measure is used for computational approaches; therefore, we aim to specify our own intuitive measure as: 

\begin{equation}
\label{eq:so}
   SO_{c_i} = \frac{A_\phi^{c_i}}{A_\sigma} 
\end{equation}
Where $A_\phi$ is defined as the occluded (not visible) surface area of the object and $A_\sigma$ stands for the total surface area of the object. 
\begin{figure}[th!]
\begin{center}
   \includegraphics[width=1.0\linewidth]{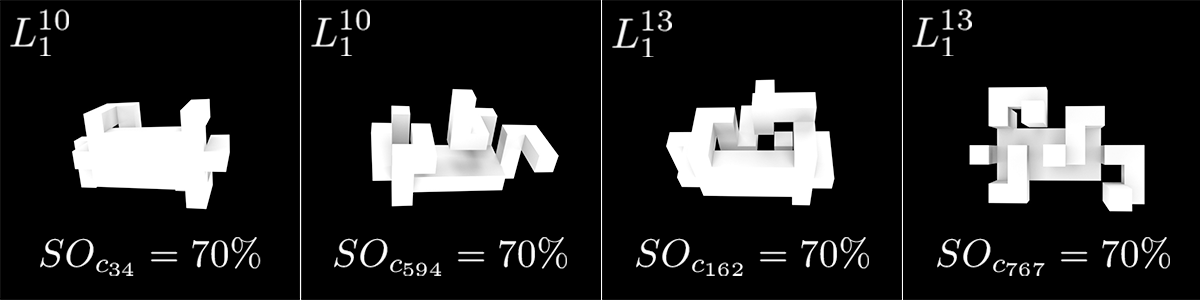}
\end{center}
   \caption{Examples of different objects (Object 10 and 13 of $L_1$) and poses causing the same amount of occlusion but different appearances.}
\label{fig:occl_vp}
\end{figure}

An object might have different views from which it causes the same amount of self-occlusion, resulting in perhaps a considerably different appearance. Figure \ref{fig:occl_vp} shows an example of two objects from two different views with the same amount of occlusion. 
Therefore, we also consider the camera's point of view with $c_i$ as the camera pose. Here, $c_i$ is defined as the camera position $c_i = (x_i, y_i, z_i)$. Therefore, we also consider the camera's point of view with $c_i$ as the camera pose. Here, $c_i$ is defined as the camera position $c_i = (x_i, y_i, z_i)$ and computed based on the Fibonacci lattice approach (see Figure \ref{fig:views}). The camera orientation is automatically set such that the object is in the centre of the viewpoint.  

\begin{figure}[th!]
\begin{center}
   \includegraphics[width=1.0\linewidth]{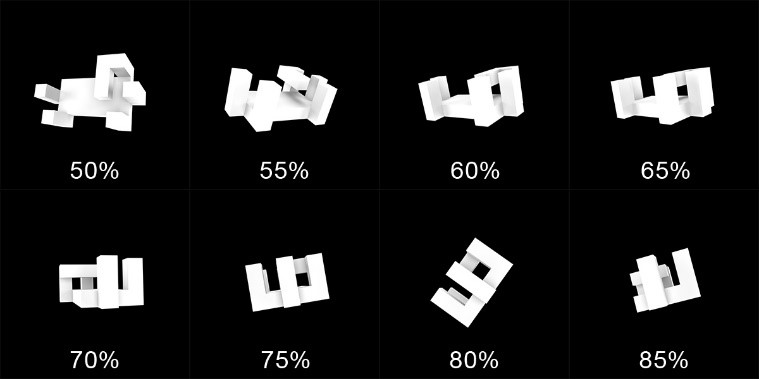}
\end{center}
   \caption{Examples of object viewpoints and their corresponding $SO_{c_i}$.}
\label{fig:occl}
\end{figure}

\begin{figure}[th]
\begin{center}
   \includegraphics[width=1.0\linewidth]{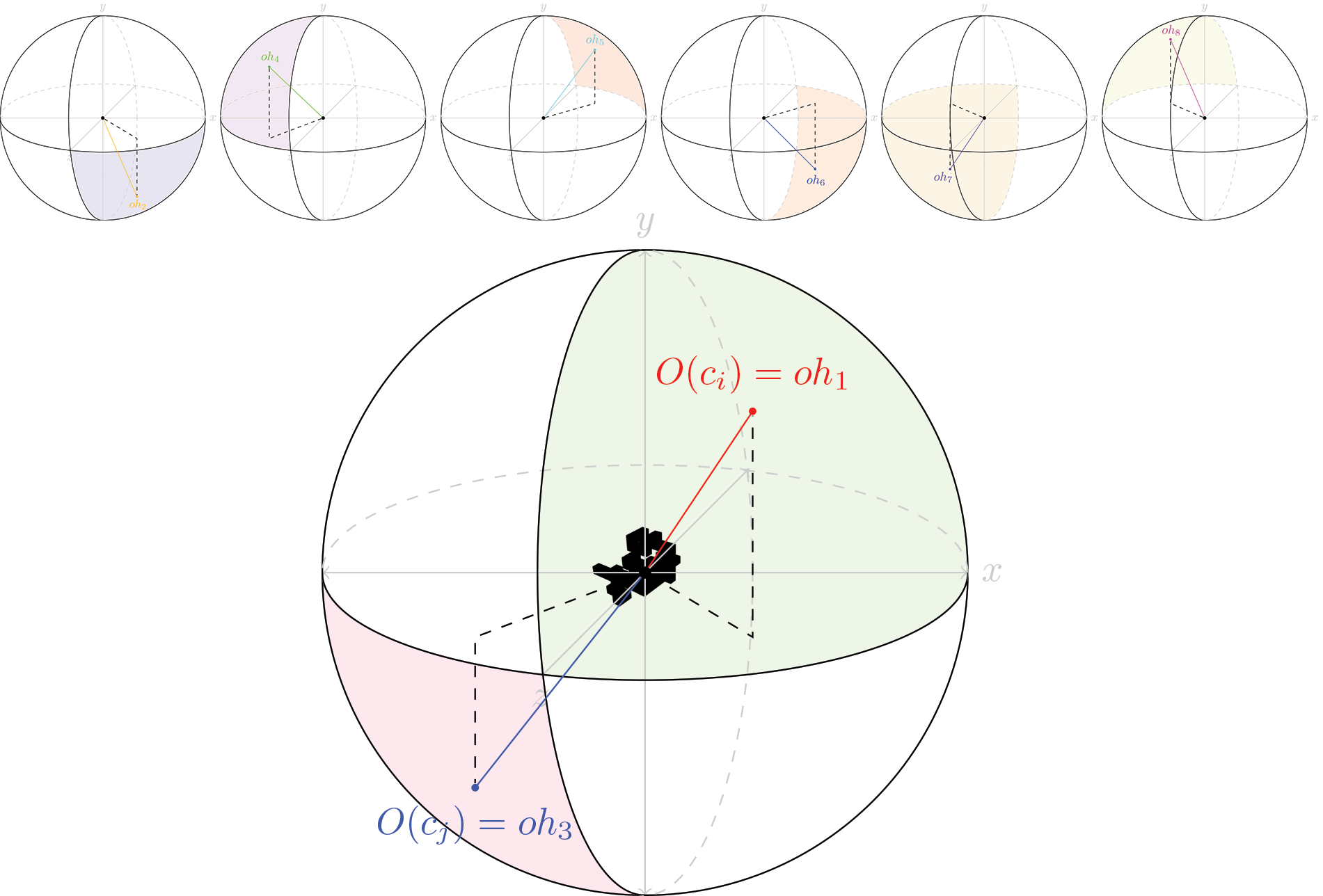}
\end{center}
   \caption{Visualization of the octahedron based projection used to map camera positions. Bottom: two example camera poses ($c_i$ and $c_j$) mapped to $oh_1$ and $oh_3$.}
\label{fig:occl_map}
\end{figure}

For evaluation purposes, we also define a function that maps a camera position ($c_i$) onto one of the eight regions of the octahedral viewing-sphere placed at the centre of an object. Figure \ref{fig:occl_map} illustrates a mapping example for two camera-positions. 

We represented the viewing sphere around an object as a spherically tiled octahedron, resulting in eight uniformly distributed triangles. To map a viewpoint $c_i$ to a tile, we perform a determinant check to see in which tile a given camera pose $c_i$ is located.
\\In our rendered data set, the self-occlusion was calculated by using the following steps:

\begin{enumerate}
   \item Iterates over all faces of the object with valid normals and calculate the ($A_\sigma$) 
   \item Subdivide the objects into many thousand elements
   \item Position the camera at a given location and pointing it at the object (see Figure \ref{fig:views}) 
   \item Select all vertices that are visible through the camera view-port
   \item Divide the object into two objects: visible and not-visible
   \item Iterate over all faces of the not-visible object with valid normals and calculate ($A_\phi$)
   \item Lastly, calculate Self-Occlusion (Equation \ref{eq:so}) 
\end{enumerate}

\begin{figure}[th!]
\begin{center}
   \includegraphics[width=1.0\linewidth]{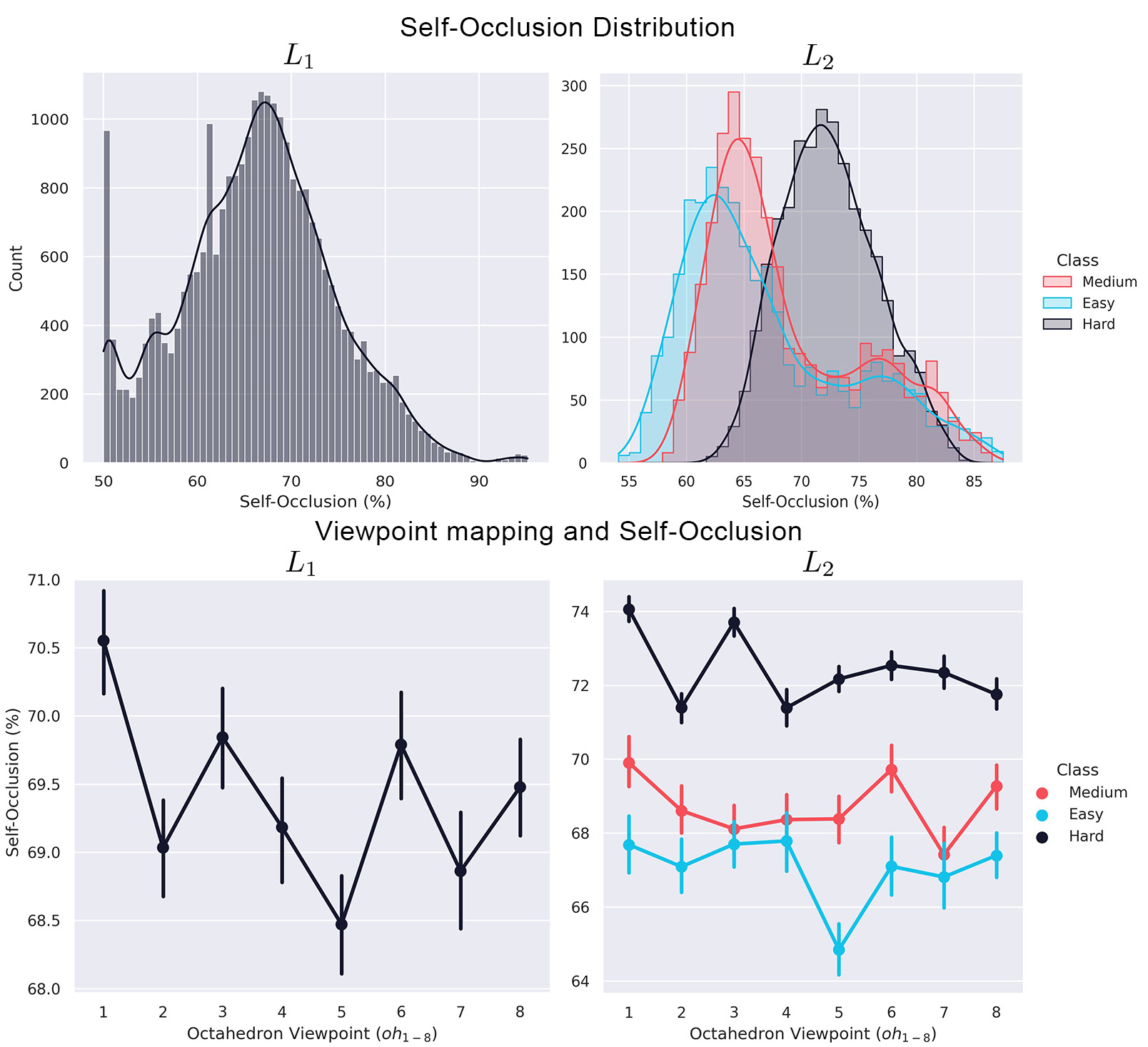}
\end{center}
   \caption{Illustration of the self-occlusion distribution for $L_1$ and $L_2$ (top), as well as the distributional relation between viewpoint mapping and self-occlusion for $L_1$ and $L_2$ (bottom).}
\label{fig:distr}
\end{figure}

Figure \ref{fig:occl} shows eight examples of the same object (object-7) from different viewing angles and sorted based on their amount of self-occlusion. As can be seen in the illustration, a single object can cast many different appearances based on the viewing angle and a significant change in the amount of what is observable of it.
\\Figure \ref{fig:distr} illustrates the self-occlusion distribution for $L_1$ and $L_2$ (top) and the distributional relation between viewpoint mapping and self-occlusion for $L_1$ and $L_2$ (bottom). Self-Occlusion for $L_1$ ranges from 49.99\% to 95.16\% with a mean at around 68\% and $L_2$ from 54.08\% to 87.5\% with a mean at 61\% (Easy), 63\% (Medium), and 71\% (Hard). The lower half of the Figure shows that different octahedron viewpoints result in varying amounts of self-occlusion. For both $L_1$ and $L_2$, a sweet-spot with the least self-occlusion is at $oh_5$, presumably resulting in the best classification result. 

\section{Baseline Evaluation}
In this Section, we discuss how well state-of-the-art classification approaches perform on \textit{TEOS}. We have chosen five deep learning models with different properties, carefully trained and evaluated them on \textit{TEOS}. 

\subsection{Model-Selection}

We have chosen Inception-V3(\cite{Szegedy_2016_CVPR}), MobileNet-V2 (\cite{sandler2018mobilenetv2}), ResNet-V2 (\cite{he2016identity}), VGG16 (\cite{simonyan2014very}) and EfficientNet (\cite{tan2019efficientnet}) as reference networks for \textit{TEOS}. Their trained version of \textit{TEOS} will be made publicly available. Table \ref{tb:cnn} shows more details about the networks in ascending order of their parameter count.
\begin{table}[ht!]
\centering
\caption{High-Level CNN Characteristics}
\begin{tabular}[t]{lcc}
\label{tb:cnn}
\textbf{CNN} & \textbf{Layers} & \textbf{Parameters (mil.)} \\
\hline
\hline
MobileNet-V2&53&3.4\\
Inception-V3&48 &24\\
ResNet-V2&152&58.4\\
EfficientNet-B7&813 &66\\
VGG16&152&138\\
\hline
\end{tabular}
\end{table}%

With \textit{TEOS}'s unique characteristics, it is not trivial to tell which CNN architecture will perform better or worse. Therefore, we chose networks with varying numbers of parameters and different numbers of layers. 

\subsection{Training-Parameter}

Besides the architecture of CNNs, a crucial element is the choice of training-parameters and so-called hyperparameters. In our case, we have looked at the input size, input noise, dropout rate, learning rate, optimization algorithm and lastly, the difference between learning from scratch and fine-tuning the networks. 
Hyperparameters such as input noise, drop rate, learning rate were determined using the hyperparameter optimizer Hyperband by \cite{li2017hyperband}. The remaining parameters were empirically determined. Table \ref{tb:param} presents the parameters used to establish the baseline of \textit{TEOS}. 
\begin{table}[ht!]
\centering
\caption{Chosen Training Parameters}
\begin{tabular}[t]{lcc}
\label{tb:param}
\textbf{Parameter} & \textbf{Value}\\
\hline
\hline
Input Sice & 224 x 224 – 800 x 800 \\ & (dependent on CNN)\\
Input Noise & Gaussian Noise of 0.1 \\
Drop Rate & 20\% \\
Learning Rate & 1e-5 \\
Optimizer & Adam Optimization \\
Learning Method & Fine Tuning \\
\hline
\end{tabular}
\end{table}%

\subsection{Training Data Preparation}
To prepare the data for training, we chose a 20\% validation split and augmented the remaining 80\% with standard data augmentation techniques (\cite{Shorten2019}). See Table \ref{tb:augm}:
\begin{table}[ht!]
\centering
\caption{Parameters for Data Augmentation}
\begin{tabular}[t]{lcc}
\label{tb:augm}
\textbf{Data Augmentation Technique} & \textbf{Value}\\
\hline
\hline
Rotation & 0-40 degrees\\
Width Shift & 0-20\%\\
Height Shift & 0-20\%\\
Zoom & 0-20\%\\
\hline
\end{tabular}
\end{table}%

\subsection{Classification Results}

\begin{figure}[th!]
\begin{center}
   \includegraphics[width=1.0\linewidth]{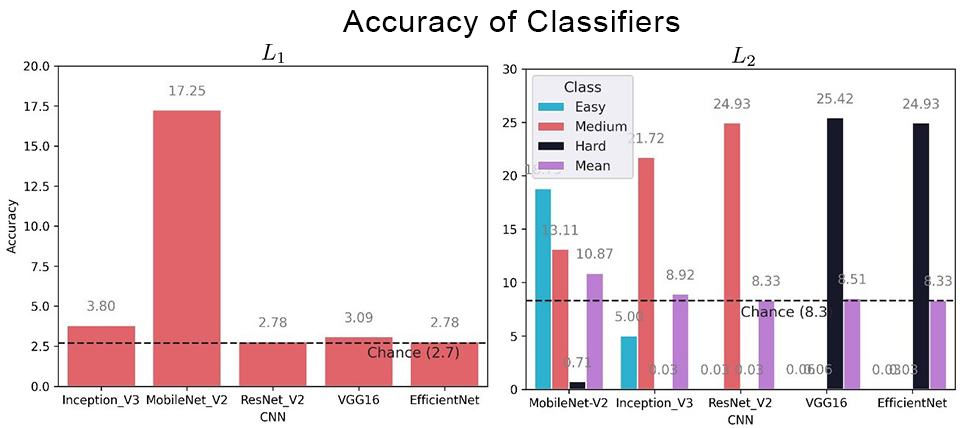}
\end{center}
   \caption{Evaluation results on $L_1$ (left) and $L_2$ (right) for five different CNNs and their accuracy across the entire datasets.}
\label{fig:acc}
\end{figure}

Our results show that MobileNet-V2 performed best across $L_1$ and $L_2$. Specifically, for $L_1$, it achieved a top-1 accuracy of 17.25\% and 10.83\% on the $L_2$ data set. See Figure \ref{fig:acc} for the classification accuracies of all networks. It seems that MobileNet-V2 is the only network that was able to learn some aspects of \textit{TEOS}, performing with a large ($L_1$) or small ($L_2$) margin above chance, whereas all other networks perform at around chance. This, perhaps, has something to do with the relatively homogeneous appearance of \textit{TEOS}, not allowing the more complex CNNs to learn from. However, this needs to be investigated further in the future.

Generally, $L_2$ is more challenging to learn for CNNs than $L_1$. Even the best performing CNN is only 2.53\% above chance, where this margin for $L_1$ was at about 14.5\%. This is explainable with the high intra-class similarity of $L_2$ – objects of one class look very similar to each other and only vary in one small detail, which might be only observable from certain viewpoints, hence will be confused with each other. 
\\The $L_1$ dataset, on the other side, has a low inter-class similarity -- the appearance of objects varies between classes.  
\\A closer look at the results of $L_2$ reveals that more extensive networks (VGG16 and EfficientNet-B7) were able to learn objects of class ``Hard'' of $L_2$; however, they could not learn ``Medium'' and ``Easy'' Objects. The smaller networks, on the other hand (MobileNet-V2 and Inception-V3), were able to learn ``Easy'' and ``Medium'' objects but not ``Hard.'' Except for MobileNet-V2, all networks have profound problems to learn the ``Easy'' Objects. See Figure \ref{fig:acc} for details.

\begin{figure}[th!]
\begin{center}
   \includegraphics[width=1.0\linewidth]{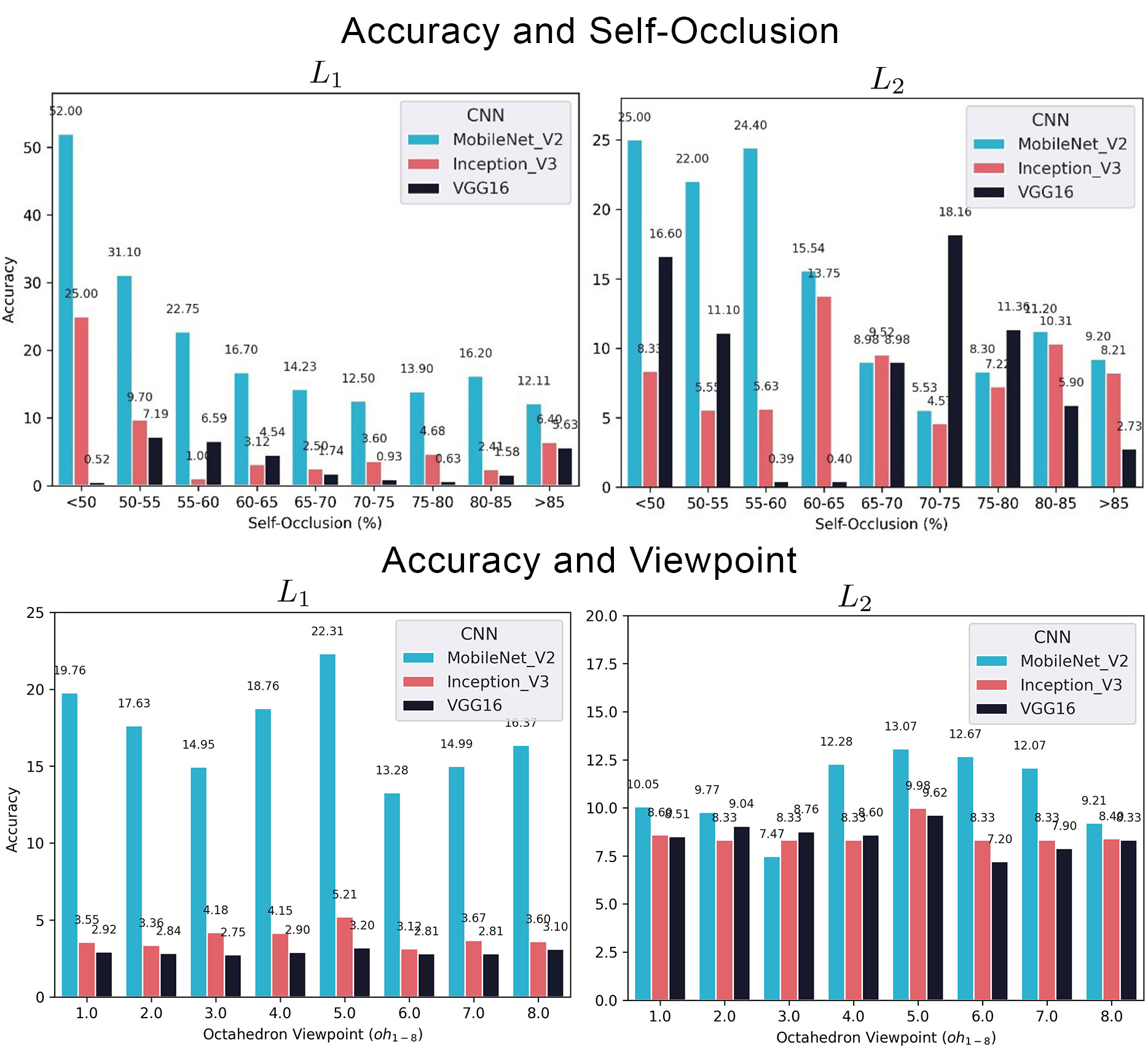}
\end{center}
   \caption{Evaluation results on $L_1$ and $L_2$ for the three top-performing CNNs. Top: Accuracy across the entire datasets with respect to self-occlusion. Bottom: Accuracy and how it is affected by the chosen viewpoint.}
\label{fig:acc_occ}
\end{figure}

Regarding the connection between classification accuracy and amount of self-occlusion, it can be generally said that the classification accuracy goes down if self-occlusion increases. We have chosen the three best-performing CNNs to analyze this connection and grouped $L_1$ and $L_2$ from 50\% to 85\% self-occlusion in 5\% intervals. $<50\%$ captures all viewpoints with a self-occlusion of less than $50\%$. $>85\%$ includes all images with more than 85\%. Furthermore, we also investigated the connection between the viewpoint mapped to an octahedral viewing-sphere and accuracy. As can be seen in the example of $L_1$ and MobileNet-V2, the viewpoint does play a vital role and can result in an increase of accuracy performance by $13.28 \hookrightarrow 22.31 = 67.99\%$. Across $L_1$ and $L_2$ the octahedral viewpoint resulting in the best performance was $oh_5$. This can be explained with that all objects share a common coordinate system and shows once more that the viewpoint matters and, even more, that an \textit{ideal viewpoint} can exist. See Figure \ref{fig:acc_occ} for details.
\\Further, even though the CNNs are trained and validated on the entire data set, their best performance can be seen at lower self-occlusion rates, which shows the vital role of self-occlusion for object classification performance. 

\section{Conclusion and Future Directions}
In this work, we have presented a novel 3D blocks world dataset that focuses on the geometric shape of 3D objects and their omnipresent challenge of self-occlusion. We have created two data sets, $L_1$ and $L_2$, including hundreds of high-resolution, realistic renderings from known camera angles. Each data set also comes with rich annotations. 
\\Further, we have presented a simple but precise measure of self-occlusion and were able to show how self-occlusion challenges the classification accuracy of state-of-the-art CNNs and the viewpoint can benefit the classification. 
\\Lastly, in our baseline evaluation, we have presented that CNNs are unable to learn \textit{TEOS}, leaving much room for future work improvements.  
\\We hope to have pathed a way to explore the relationship between object classification, viewpoint, and self-occlusion with this work. Specifically, we hope that \textit{TEOS} is useful for research in the realm of active vision – to plan and reason for the next-best-view seems to be crucial to increase object classification performance. 

\section{Acknowledgements}
This research was supported by grants to the senior author (JKT) from the following sources: Air Force Office of Scientific Research USA (FA9550-18-1-0054), The Canada Research Chairs Program (950- 231659) and Natural Sciences and Engineering Research Council of Canada (RGPIN-2016-05352). 
The funders had no role in study design, data collection and analysis, decision to publish, or preparation of the manuscript.

{\small
\bibliographystyle{apalike}
\bibliography{library}
}

\end{document}